\documentclass[lettersize,journal]{IEEEtran}
\usepackage{amsmath,amsfonts}
\usepackage{algorithmic}
\usepackage{algorithm}
\usepackage{array}
\usepackage[caption=false,font=normalsize,labelfont=sf,textfont=sf]{subfig}
\usepackage{textcomp}
\usepackage{stfloats}
\usepackage{url}
\usepackage{verbatim}
\usepackage{graphicx}
\usepackage{cite}
\usepackage{multirow}
\usepackage{caption}  
\usepackage{bm}
\usepackage{diagbox}      
\usepackage{xcolor}       
\usepackage{booktabs}     
\usepackage{colortbl}     
\usepackage{float} 
\usepackage{hyperref}

\begin{document}



\title{TacRefineNet: Goal-Conditioned Tactile Grasp Refinement for Edge-Prominent Objects
}

\author{
    Shuaijun Wang, Haoran Zhou, Diyun Xiang, Yangwei You\\
    
    \thanks{All authors are with Xiaomi Robotics. Corresponding author: Shuaijun Wang, email:wukongwoong@gmail.com.
}}



\IEEEaftertitletext{
    \centering
    \includegraphics[width=\textwidth]{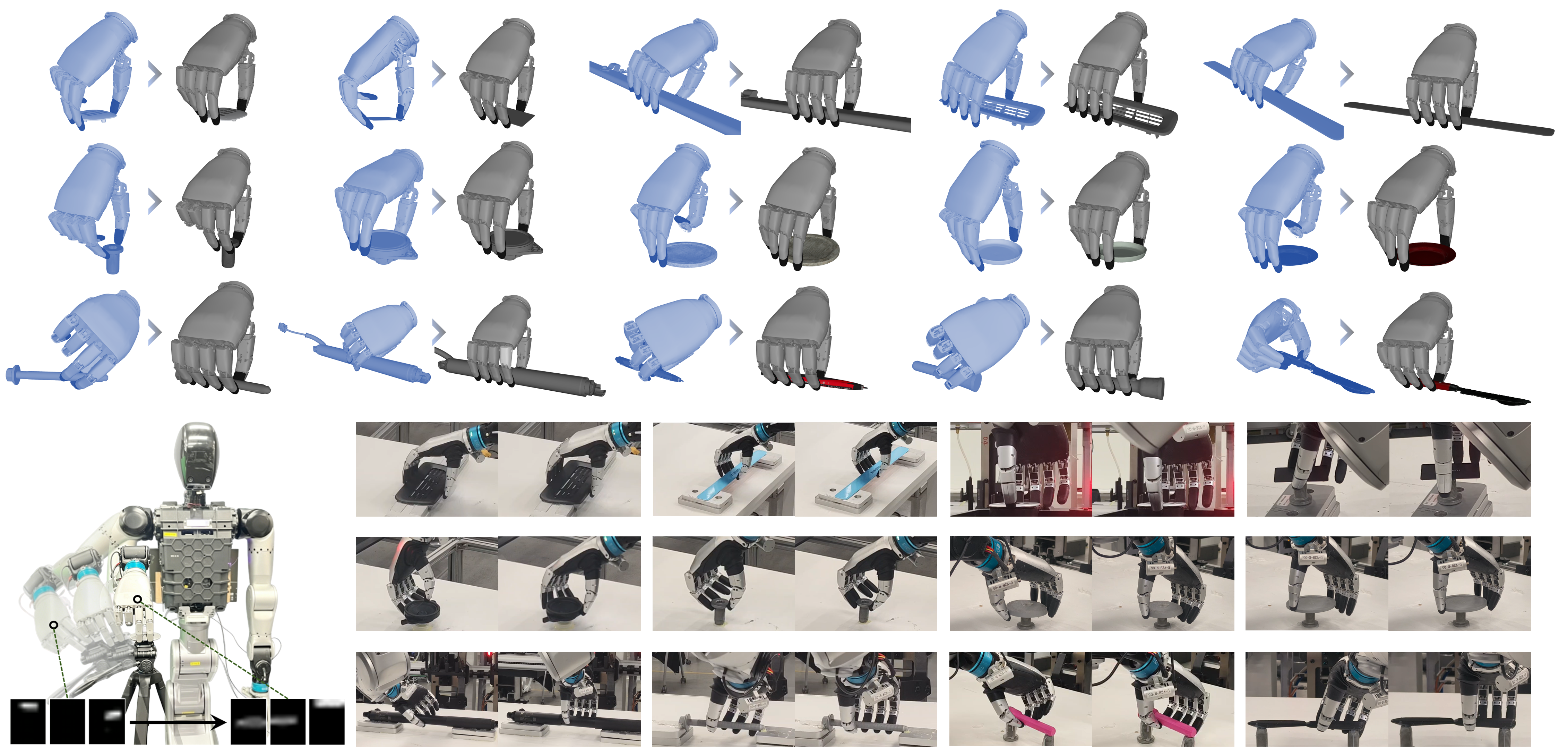}
    \captionof{figure}{Zero-shot sim-to-real transfer for multi-finger tactile-guided grasp refinement across three object categories: rectangular plates, discoid objects, and slender rods.}
    \label{fig:teaser}
    \vspace{1em}
}

\maketitle

\begin{abstract}
Accurate final grasp alignment remains challenging for edge-prominent objects such as thin plates, discs, and rods, whose sparse contacts are easily occluded and poorly resolved by depth sensing. We present TacRefineNet, a tactile-only, goal-conditioned framework for local refinement along tactilely observable pose dimensions. Given current and target multi-finger tactile images and their corresponding hand-joint configurations, a Siamese policy network directly predicts corrective wrist pose increments. The hand iteratively opens, moves, and regrasps, forming an external-dexterity tactile servoing loop. Cross-combination training pairs current and target samples, allowing targets within the sampled pose range to be specified without retraining. We collect 156,007 simulated samples from 15 plates, discs, and rods and train the policy entirely in MuJoCo before zero-shot deployment to an 11-DoF five-fingered hand with piezoresistive sensors. On seen objects, the real system achieves 80.7\% and 59.3\% success under the $10^\circ$/10\,mm criterion for fixed and random targets, respectively; after five steps, the mean errors are approximately 5.2\,mm and $3.5^\circ$. Experiments further show continuous correction under long-horizon perturbations and limited within-category transfer to unseen objects, with reduced performance for symmetric or weakly discriminative contacts.
Project website is available at \href{https://sites.google.com/view/tacrefinenet}{https://sites.google.com/view/tacrefinenet}.
\end{abstract}


\section{Introduction}

Robotic fine grasping remains a long-standing challenge in robotics~\cite{newbury2023deep,zhou2025dydexhandover,wang2017generic,wang2023learning}. While conventional pipelines decompose grasping into approaching~\cite{morales2006integrated}, planning~\cite{zhong2025dexgrasp}, and execution~\cite{wang2022learning}, and recent end-to-end Vision-Language-Action (VLA) models learn policies directly from high-dimensional visual inputs~\cite{wen2025tinyvla,zitkovich2023rt,zhao2023learning,chi2023diffusion,wang2024learning}, accurate execution in the final stage remains brittle. In long-horizon manipulation, small perception and control errors accumulate, and the resulting pose deviation during grasp execution can cascade into failures in downstream tasks such as insertion and assembly. This challenge is particularly acute for \textit{edge-prominent objects}, namely objects whose stable grasping and downstream use depend on thin, flat, circular, or elongated contact edges. Examples include thin plates, discs, and slender rods, where contact is sparse and even minor misalignment can destabilize the grasp.

For edge-prominent objects, visual feedback can be unreliable precisely at the moment when precision matters most: contacts often occur at thin edges with \emph{sparse visible cues}, and the contact region is frequently \emph{occluded} by fingers/hand. In addition, depth sensing can be noisy or blurred near reflective or thin structures, making it difficult to infer minor pose errors from pixels alone. These failure modes motivate using touch as a primary signal for robotic grasping refinement.

Tactile sensing directly measures the contact interface and therefore provides information complementary to vision~\cite{luo2025tactile,huang20243d,9681179}. We propose \textbf{TacRefineNet}, a goal-conditioned policy for tactile-only grasp refinement that iteratively refines the local pose of a grasped object toward a user-specified target using multi-finger tactile observations. In this work, the target is represented by a reference tactile observation, and the policy operates within the local pose range covered by the collected data. The proposed framework is evaluated on three representative classes of edge-prominent objects: thin rectangular plates, flat discoid objects, and slender rods.

We integrate custom piezoresistive tactile sensors into an 11-DoF five-fingered hand. Each fingertip provides a high-resolution taxel array, which is treated as an image-like tactile input to the policy. Given the underactuated kinematics and limited in-hand controllability, we exploit \textbf{external dexterity}: instead of relying solely on finger gaiting to move the object in hand, the system changes the relative object--hand pose by moving the wrist and regrasping. Specifically, the hand grasps, observes multi-finger tactile feedback, predicts a corrective wrist motion, and regrasps at the updated wrist pose until the tactile observation matches the target.

To train the policy, we construct a large-scale simulated dataset in a physics-based MuJoCo tactile simulator containing 15 objects (five per geometry category). TacRefineNet adopts a Siamese architecture with weight-shared encoders for the current and target tactile observations, which are fused with the corresponding current and target proprioceptive states to predict wrist pose increments. The policy is trained using a cross-combination strategy, where pairs of current and target tactile--proprioceptive samples are formed to supervise wrist pose increments. During both training and deployment, target poses can be selected within the pose range covered by the dataset. 

At deployment, any target tactile observation within this range can be specified without requiring additional training. We evaluate the proposed method in both simulation and real-world experiments on the training objects as well as previously unseen objects from the same geometry category, including objects with significantly different tactile edge characteristics. Experimental results demonstrate millimeter-level pose refinement accuracy on seen objects and show that the learned tactile policy can transfer to unseen objects within the same geometry class, although with reduced accuracy when tactile contact patterns become less discriminative. In addition, extensive ablation studies analyze the impact of the network architecture, proprioceptive input, and the number of tactile sensing fingers, providing further insights into the key design choices of the proposed framework.

In summary, our contributions are:
\begin{itemize}
    \item \textbf{A goal-conditioned formulation for tactile-based grasp refinement.}
    We formulate fine grasp execution as a tactile servoing problem that iteratively aligns the current multi-finger tactile observation with a target tactile observation through an external-dexterity regrasp loop, eliminating the need for vision, object models, or target-specific retraining.

    \item \textbf{A Siamese multi-finger tactile policy for relative grasp-pose correction.}
    We develop TacRefineNet to jointly encode paired current and target tactile observations, aggregate contact information across fingers, and fuse the corresponding proprioceptive states. Current--target cross-attention directly regresses corrective wrist-pose increments along geometry-dependent tactile-observable DoFs, avoiding intermediate absolute-pose estimation and supporting different targets within the sampled local pose range.

    \item \textbf{Extensive sim-to-real validation.}
    We train the policy entirely in a physics-based tactile simulator and transfer it zero-shot to a real robotic hand. Experiments on seen objects and within-category unseen objects with novel tactile features, together with robustness evaluations and ablation studies, quantify local refinement performance and clarify the capabilities and limitations of the proposed framework.
\end{itemize}

\section{Related Works}

\subsection{Tactile-Proprioceptive-Action Datasets for Fine Grasping}
Data-driven tactile grasping and manipulation have gained increasing attention in recent years\cite{li2014learning,su2024sim2real,chen2025fbi}. The emergence of large-scale multimodal datasets has supported the development of deep learning-based methods. Wan et al. introduced VinT-6D, a multimodal in-hand manipulation dataset containing 2 million simulation samples and 100k real-world samples, combining vision, tactile, and proprioceptive information~\cite{wan2023vint}. This is a significant improvement compared to earlier datasets such as those by Wen et al.~\cite{wen2020tactilegrasp}, which only contained tens of thousands of grasps, or the VITA dataset proposed by Dikhale et al.~\cite{dikhale2022vita}. In simulation tools, Leins et al. proposed HydroElasticTouch, a MuJoCo plugin that realistically simulates resistive tactile sensors for high-fidelity tactile signal generation~\cite{leins2025hydroelastictouch}. Additionally, Adeniji et al.~\cite{adeniji2023tactileglove} collected human tactile-proprioception demonstrations via a sensorized glove and used a Transformer-based policy to retarget actions to a robotic gripper. In contrast to datasets aimed at perception or vision-based manipulation, our dataset focuses specifically on learning the mapping between tactile, proprioception, and control actions. We establish a unified simulation training pipeline, collect fingertip tactile data, and pair it with ground-truth robot control actions for training.

\subsection{Tactile-Based Grasp Refinement}
Tactile sensing has been widely applied in improving grasp performance\cite{su2024sim2real}. Prior research has used tactile feedback for grasp quality prediction, slip detection, and object pose estimation \cite{su2023soft}. For instance, Hogan et al.~\cite{hogan2020tactile} learned a tactile quality metric from GelSight images and applied local regrasping by perturbing tactile inputs. Mao et al.~\cite{mao2024tactilefeatures} trained autoencoders to extract latent tactile features for precise pinch grasps, enabling reactive regrasping when objects slip. These methods mentioned above enhance grasp robustness\cite{zhao2024tactile}. However, they are generally limited to evaluating grasp quality or estimating local poses. In contrast, our method uses multi-finger tactile images to directly control the robot in an end-to-end fashion. It achieves fine grasping adjustment for grasping pose deviations using only fingertip tactile feedback, without any prior object model or external perception, enabling continuous tactile-driven refinement to reach the target pose.

\begin{figure*}[h!t]
\centering
\includegraphics[width=7in]{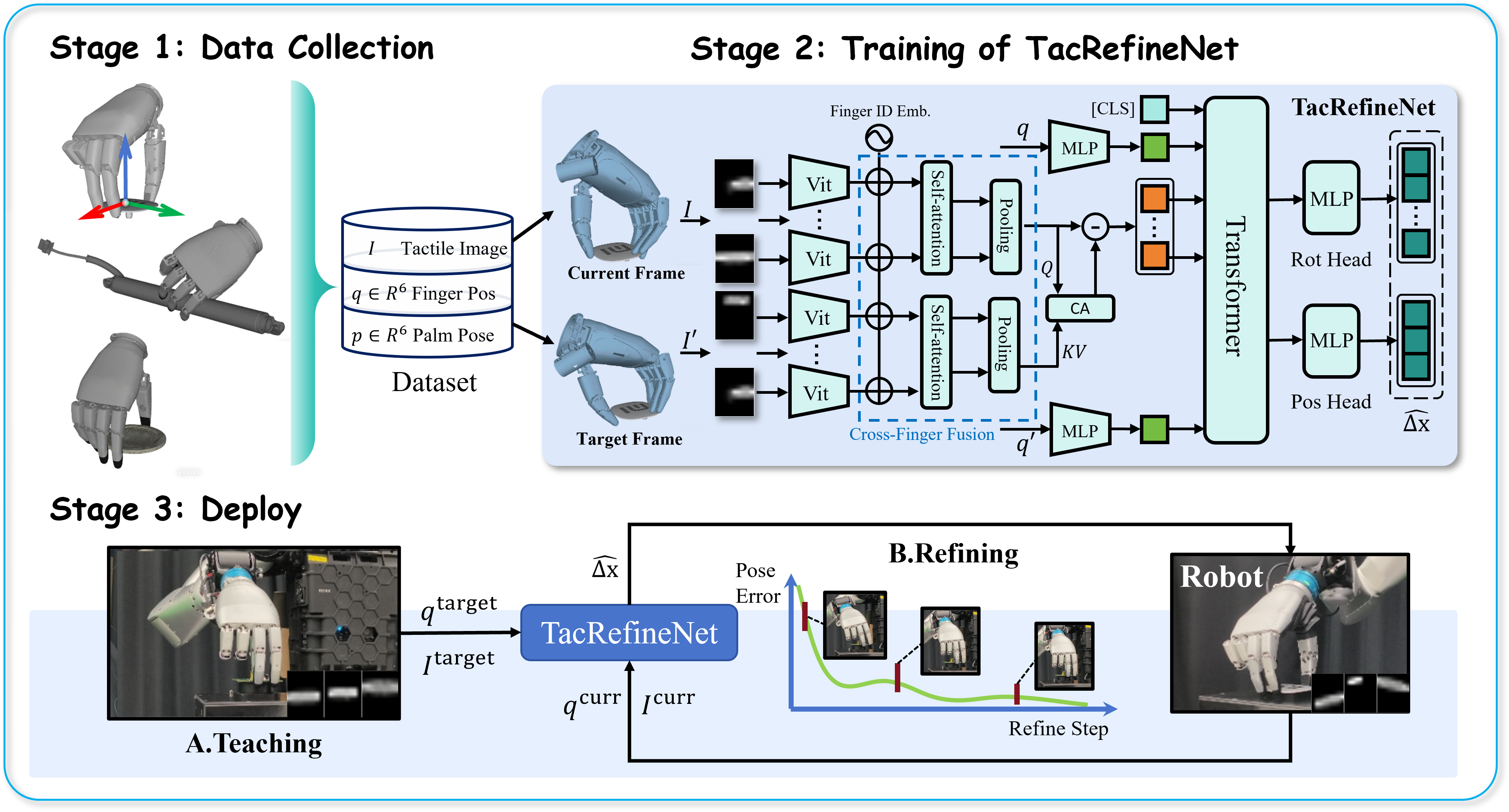}
\caption{\textbf{Overview of TacRefineNet.} (a) Simulated data collection. (b) Siamese network architecture and cross-combination training: a pair
     of tactile frames is drawn as the target frame and current frame, and the network regresses the wrist-pose increment that aligns
     the current frame to the target. (c) Real-world deployment, where a demonstration provides the target frame and the hand iteratively regrasps
     to refine the grasping pose.}
\label{framework}
\end{figure*}

\subsection{Tactile-Driven Policy Learning}
Learning-based control for dexterous hands using tactile information has made rapid progress\cite{chen2025fbi,su2024sim2real,zhao2024tactile,liang2023visuo}. Yang et al. proposed TacGNN~\cite{yang2023tacgnn}, which learns blind manipulation from multi-finger tactile signals using graph neural networks and reinforcement learning. Pitz et al.~\cite{pitz2024tactilegoal} introduced goal-conditioned tactile RL for object alignment with proprioception and tactile input under known object geometry. While these methods demonstrate the potential of tactile sensing, they either rely on object priors or lack generalizable fine-grained control. In contrast, our approach employs a Siamese tactile policy network with shared encoders for the current and target observations, directly regressing pose increments from paired tactile images and their corresponding proprioceptive states. This design enables goal-driven regrasping without vision or target-specific retraining. TacRefineNet achieves millimeter-level accuracy for target poses using low-cost tactile sensors in the evaluated local refinement setting.

\section{Method}

\subsection{Method Overview}
We propose \textbf{TacRefineNet}, a tactile-driven framework for local grasp refinement at the final stage of robotic grasping. An upstream perception and planning system first provides a coarse grasp. Residual errors may remain because of uncertainty in perception, planning, and execution, particularly for edge-prominent objects with sparse or occluded contact features. TacRefineNet uses fingertip tactile feedback to correct these errors after contact is established.

During each refinement step, the multi-finger hand grasps the object and records the current fingertip tactile observations. The policy uses these observations to predict a corrective wrist-pose increment. The hand then reopens, moves to the updated wrist pose, and regrasps the object. Repeating this closed-loop process progressively adjusts the grasp toward the target configuration.

The goal is represented by a reference tactile observation and its corresponding hand-joint configuration within the pose range covered by the dataset, such as a pair recorded from a human demonstration. These reference signals are provided to the policy as the target tactile and proprioceptive inputs. During refinement, the method does not require external vision or an object model. 

As shown in Fig.~\ref{framework}, the complete framework consists of simulated data collection, policy training, and real-world deployment. The simulator provides tactile--proprioception--action samples, and cross-combination training forms current--target observation pairs with their corresponding wrist-pose increments. The resulting policy is then applied iteratively to refine an initial grasp toward a specified target grasp.

Formally, the TacRefineNet policy $\pi$ is defined in Eq.~\ref{equ_policy} as
\begin{equation}
\Delta \mathbf{x} = \pi\left( \left\{ \mathbf{I}_i^{\text{curr}}, \mathbf{I}_i^{\text{target}} \right\}_{i=1}^N, \mathbf{q}^{\text{curr}}, \mathbf{q}^{\text{target}} \right)
\label{equ_policy}
\end{equation}
where $\mathbf{I}_i^{\text{curr}}$ and $\mathbf{I}_i^{\text{target}}$ are the current and target tactile images from the $i$-th fingertip, respectively; $N$ is the number of tactile fingers; and $\mathbf{q}^{\text{curr}}$ and $\mathbf{q}^{\text{target}}$ are the current and target hand-joint configurations. The output $\Delta\mathbf{x}$ represents the corrective 6-DoF wrist-pose increment. For network regression, it is encoded as the concatenated vector $\Delta\mathbf{x}=[R_9,p_3]\in\mathbb{R}^{12}$, where $R_9\in\mathbb{R}^9$ is the vectorized $3\times3$ relative rotation matrix and $p_3\in\mathbb{R}^3$ is the relative translation vector.

\subsection{Tactile Sensor and Its Simulation}
\begin{figure*}[h!t]
\centering
\includegraphics[width=7.0in]{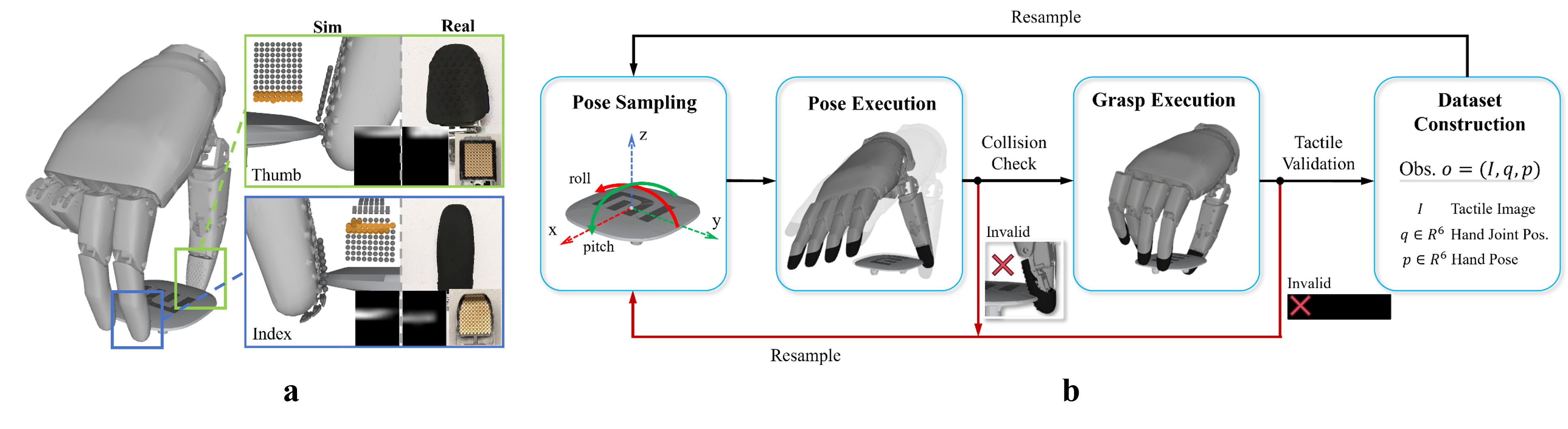}
\caption{(a) Tactile sensor and its simulation; (b) Dataset collection pipeline.}
\label{data-collection}
\end{figure*}

We use a piezoresistive tactile sensor array integrated into an 11-DoF dexterous hand. As shown in Fig.~\ref{data-collection} (a), each fingertip sensor consists of an $11 \times 9$ taxel grid, where each taxel measures normal contact force. The physical spacing between the taxels on the real tactile fingertip is approximately 1.1 mm. The raw taxel outputs are transformed into tactile images.

We develop a physics-based tactile simulation in MuJoCo for scalable data generation. The sensor is modeled as spherical contact points matching the real device's resolution and spacing; to mimic the rubber surface, each point is coupled to the sensor by an elastic joint with a contact sensor, so points deflect on contact and press back as force grows, and the aggregated responses form the tactile image.

\begin{figure}[t]
    \centering
    \includegraphics[width=\linewidth]{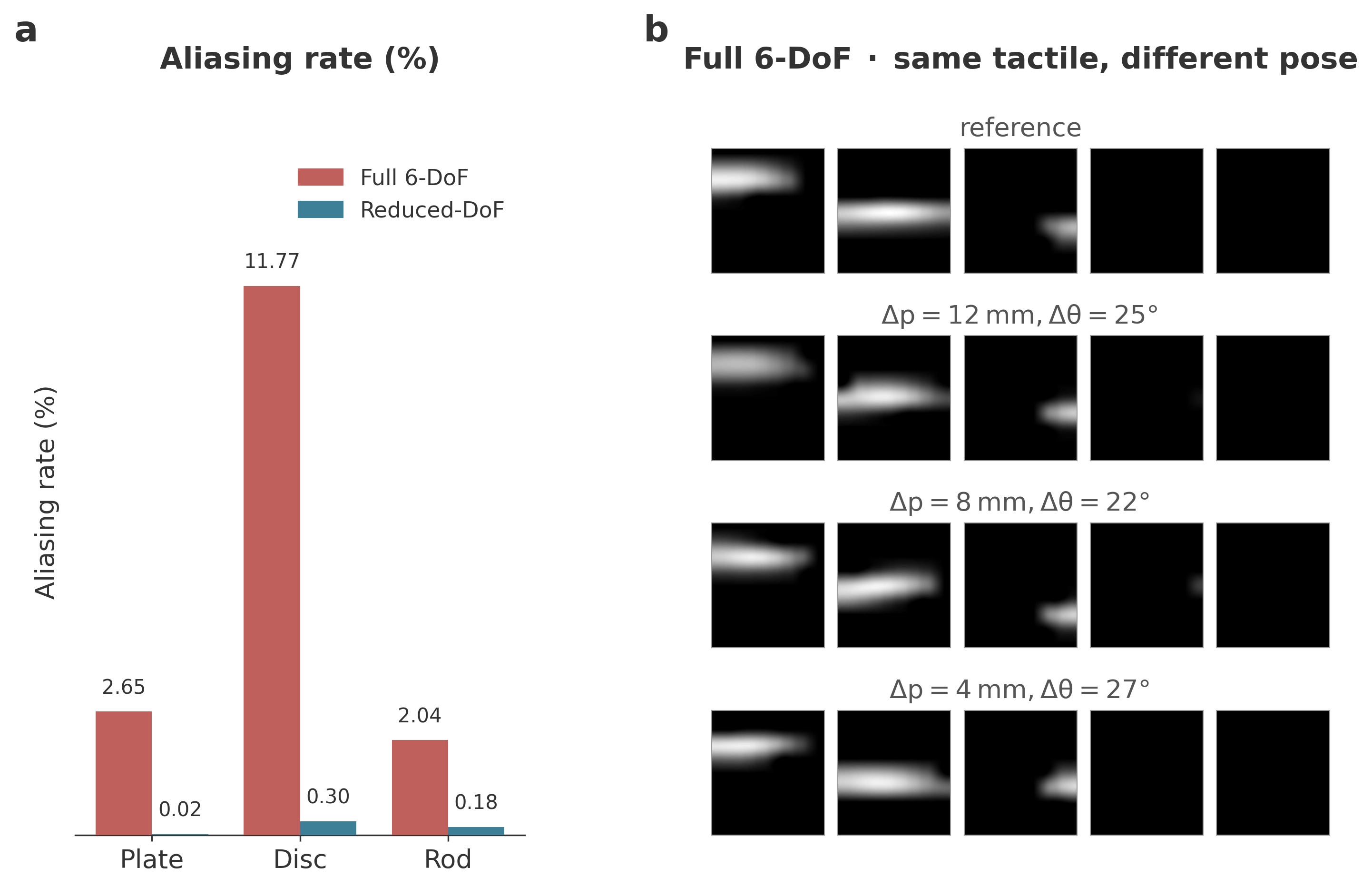}
    \caption{Tactile observation aliasing analysis and tactile image comparison across pose dimensions.}
    \label{fig:aliasing}
\end{figure}

\subsection{Simulation Data Collection and Tactile Aliasing Analysis}
Based on the tactile systems described above, we construct a simulated dataset, with the data collection process shown in Fig. \ref{data-collection}. The dataset spans 15 objects evenly distributed across the three geometry classes---thin rectangular plates, flat discoid objects, and slender rods (Fig.~\ref{fig:objects}).

In simulation, we generate the dataset by systematically sampling the hand pose along a \emph{reduced} set of pose DoFs selected per object geometry (detailed below), while holding the remaining DoFs fixed.
At every sampled pose, the hand's end-effector is set directly to the target wrist pose with the fingers fully open. We then run a penetration check and discard any pose in which the open hand already intersects the object before closing. 
Otherwise, the hand closes under MuJoCo position control, actuating the six driven joints from an open to a fixed closed configuration, with the passive PIP/DIP joints following through a fixed polynomial coupling. Rather than terminating on a force threshold, each finger halts where contact physically arrests the commanded closure. Samples that yield no valid fingertip contact after closure are rejected and re-sampled (Fig.~\ref{data-collection}). 
Once a valid grasp is obtained, we record the wrist pose $\mathbf{P} \in \mathbb{R}^6$ (measured after finger closure), the joint configuration $\mathbf{q}$, and the tactile observations $\mathbf{I}$. All samples are stored in the simulation dataset.

Touch only reveals pose changes that alter the fingertip contact distribution;
motions leaving the contact geometry invariant produce no signal. In our grasp
the fingers contact the object along its principal axis with the contact normal
along the palm ($z$) axis, so a $z$ translation changes contact depth and an
in-plane tilt (roll) induces a depth gradient---both resolvable. Axial spin
(pitch), axial slide ($y$), and in-plane motions (lateral slide $x$, yaw)
instead leave the contact essentially unchanged, so they are tactilely
unobservable and regressing them injects noise. We therefore refine only the
observable DoFs, adapting to geometry: $\{z,\mathrm{roll}\}$ for a rod,
$\{z,\mathrm{roll},\mathrm{pitch}\}$ for a plate, and $\{y,z,\mathrm{pitch}\}$
for a disc. Fig.~\ref{fig:aliasing} supports this empirically.

\subsection{Learning Fine-Grained Grasp Adjustment Policy}

\subsubsection{TacRefineNet Siamese Architecture}
The proposed network first employs a Vision Transformer (ViT) encoder to extract spatial representations from the tactile image of each fingertip independently. To preserve finger identity, learnable positional embeddings are added to the fingertip features before a self-attention module aggregates information across multiple fingers. The fused tactile tokens are then compressed through a pooling layer to obtain a compact multi-finger tactile representation.
A Siamese network architecture with shared encoder weights is adopted for the current and target tactile observations. At the feature fusion stage, the current tactile features serve as queries to attend to the target tactile features, while the target branch provides conditional information through cross-attention. The resulting attended features are added back to the current tactile tokens, yielding a unified representation that hierarchically integrates single-finger perception, multi-finger interactions, and the correspondence between the current and target tactile states.
Finally, the fused tactile representation is concatenated with the current and target proprioceptive joint states and fed into a Transformer. Two independent MLP prediction heads are employed to regress the translational and rotational pose increments, respectively.
The final network architecture used in this paper is illustrated in Fig. \ref{framework}.

\subsubsection{Policy Training}
\label{policy_AB}
We train TacRefineNet exclusively on simulated data and deploy the resulting policy on the real system in a zero-shot manner, without retraining for new targets. The network takes as input the current and target fingertip tactile images during contact, together with the corresponding current and target hand-joint states, and outputs the wrist-pose increment.

To supervise target poses within the dataset pose range without additional training, we adopt a cross-combination paradigm: current and target tactile images are randomly paired from the dataset, and the corresponding hand poses are subtracted to form the ground-truth pose increment. Constructing an $N \times N$ set of such pairwise combinations encourages the policy to generalize across diverse target poses. The network is optimized using the Mean Squared Error (MSE) loss between the predicted and ground-truth wrist pose increments (Equation~\ref{equ_loss}).
\begin{equation}
\mathcal{L}_{\text{MSE}} = \frac{1}{K} \sum_{i=1}^{K} \left\| \Delta \hat{\mathbf{x}}_i - \Delta \mathbf{x}_i \right\|^2,
\label{equ_loss}
\end{equation}
where $\Delta \hat{\mathbf{x}}_i$ is the predicted pose increment, $\Delta \mathbf{x}_i$ is the ground-truth pose increment, and $K$ is the batch size. The rotational increment is parameterized as a 9D rotation matrix and symmetrically orthogonalized via SVD, rather than represented minimally in 3D. The MSE is computed in the normalized action space over the concatenated $[\,R_{9},\,p_{3}\,]$ vector and minimized via backpropagation until convergence.
Moreover, to improve policy robustness and facilitate Sim-to-Real transfer, we employ several data augmentation techniques during training. The augmentation settings are summarized in Table~\ref{table:augment}.
\begin{table}[!htbp]
\centering
\caption{Data augmentation stetup.}
\label{table:augment}
{\begin{tabular}{@{\extracolsep\fill}lccc}%
    \toprule
    Parameter & Probability Distribution  & Operation \\
    \midrule
    Tactile Image     & \( \sim\mu \)(0.8, 1.2)   & Scaling\\
    Joint Position    & \( \sim\mu \)(-0.04, 0.04)   & Additive\\
    \bottomrule
\end{tabular}}
\end{table}

At inference, the system receives a selected target tactile image within the dataset pose range and the current tactile image from the latest grasp. Leveraging the learned mapping from tactile perception to action, the policy network outputs a wrist pose increment.

\section{Experiments}

\subsection{Experimental Setup}
\begin{figure}[h!t]
\centering
\includegraphics[width=3.5in]{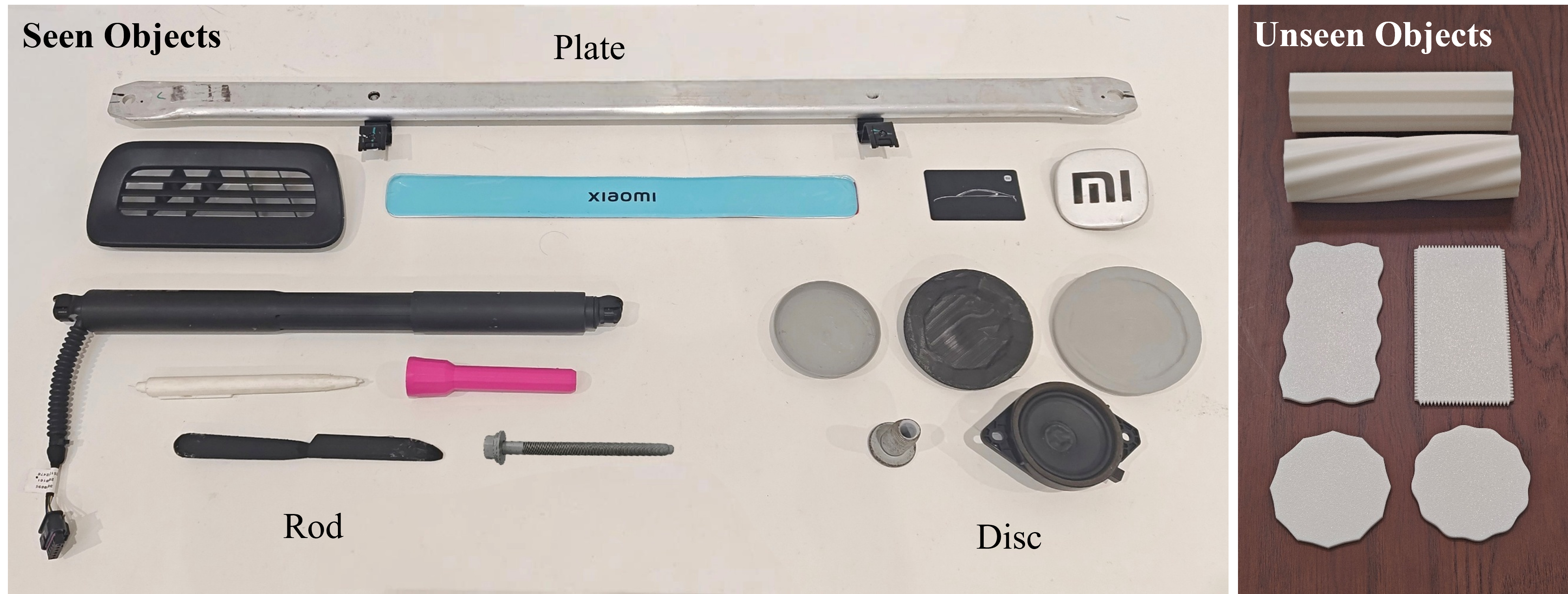}
\caption{Experimental objects. Left: five training objects per category (plates, discs, and rods). Right: two unseen objects per category for generalization evaluation.}
\label{fig:objects}
\end{figure}
\subsubsection{Training Dataset}

The training dataset consists of tactile observations collected in simulation from 15 objects, including five thin rectangular plates, five flat discs, and five slender rods, as shown in Fig.~\ref{fig:objects}. For each object category, data are collected by sampling only the pose DoFs that are observable through tactile sensing, rather than the full 6-DoF pose. In total, the dataset contains 156,007 samples. For each object, the data are randomly split into $90\%$ training and $10\%$ validation sets. During training, diverse current--target tactile pairs are generated online using the cross-combination strategy, substantially increasing the diversity of refinement tasks without additional data collection.

\subsubsection{Hardware Platform}

Simulation experiments are conducted in the MuJoCo tactile simulation environment. Real-world experiments are performed on our dual-arm robotic platform equipped with the proposed 11-DoF dexterous hand. Each fingertip integrates an $11\times9$ piezoresistive tactile array that outputs discrete pressure values in the range of $0$--$255$. During deployment on an NVIDIA Jetson Orin, tactile preprocessing takes approximately $90$\,ms and network inference requires approximately $10$\,ms, yielding a closed-loop control frequency of approximately $10$\,Hz.

\subsection{Evaluation Protocol and Metrics}

The object remains rigidly fixed throughout each experiment. Each episode begins with a teaching stage, in which the hand is placed at the desired grasp to record the target tactile observations, target hand-joint configuration, and corresponding reference wrist pose. The hand is then moved to a random initial pose within the dataset range, and the policy performs up to five iterative wrist-pose updates. Depending on the evaluation setting, the target is either the demonstrated grasp pose or a pose randomly sampled from the dataset.

The current wrist pose is obtained through forward kinematics and compared with the reference wrist pose. Because the object is stationary during the refinement progress, this difference is also the relative hand-object pose error. Each object is evaluated in 100 independent simulation trials and 10 real-world trials. Results are averaged over the five seen objects and two unseen objects in each category.

We measure both position and orientation errors. The position error is defined in Eq.~\ref{equ_pos_error} as
\begin{equation}
Err_{pos}=\left\|\mathbf{p}\right\|_2,
\label{equ_pos_error}
\end{equation}
where $\mathbf{p}$ is the translation difference between the current and reference wrist poses, and $\|\cdot\|_2$ denotes the Euclidean norm. The orientation error is the quaternion geodesic distance defined in Eq.~\ref{equ_rot_error}:
\begin{equation}
Err_{rot}=2\cos^{-1}\left(\left|\left\langle \mathbf{q}_1,\mathbf{q}_2 \right\rangle\right|\right),
\label{equ_rot_error}
\end{equation}
where $\mathbf{q}_1$ and $\mathbf{q}_2$ are the unit quaternions of the current and reference wrist orientations, respectively. 


\begin{figure}[t]
    \centering
    \includegraphics[width=1.0\linewidth]{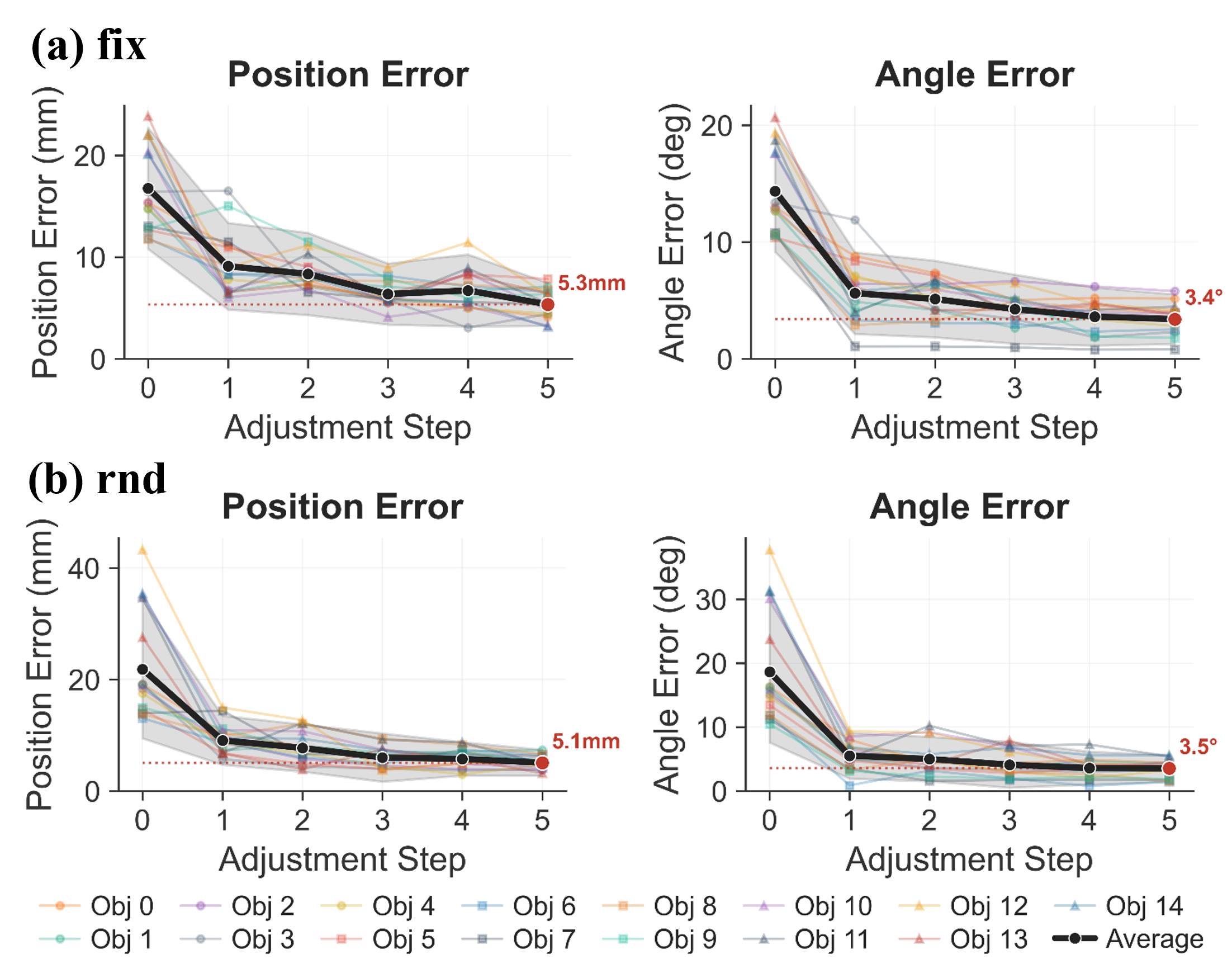}
    \caption{Real-world refinement errors for \emph{seen} objects under fixed (top) and random (bottom) targets. Left: position; right: orientation. Black curves and shading denote the mean and standard deviation.}
    \label{fig:real_seen}
\end{figure}

\begin{figure}[t]
    \centering
    \includegraphics[width=1.0\linewidth]{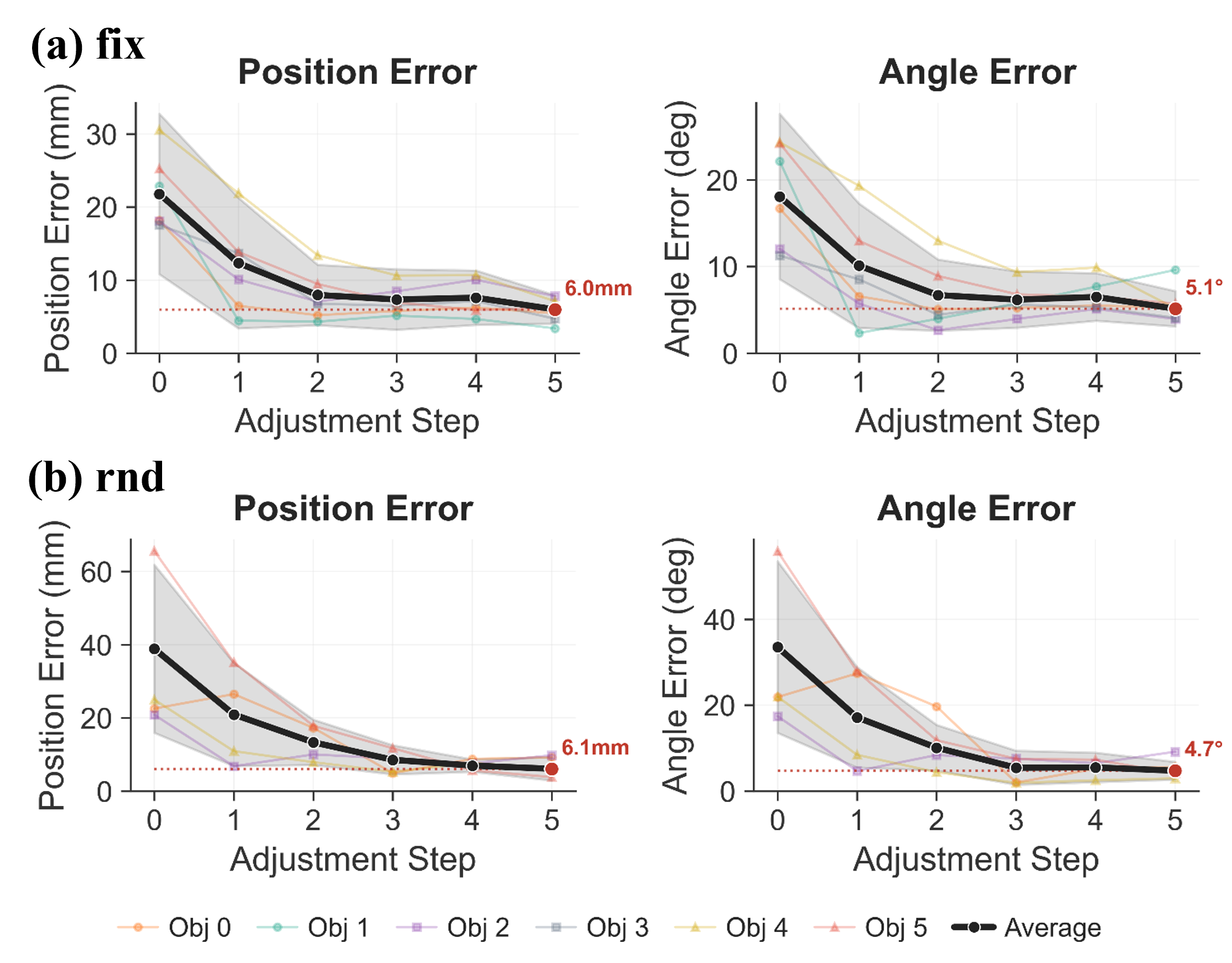}
    \caption{Real-world refinement errors for \emph{unseen} objects under fixed (top) and random (bottom) targets. Left: position; right: orientation. Black curves and shading denote the mean and standard deviation.}
    \label{fig:real_unseen}
\end{figure}

\begin{table}[!htbp]
  \centering
  \caption{\textbf{Refinement performance under the fixed-target setting.}}
  \label{tab:results_fix}
  \small
  \renewcommand{\arraystretch}{1.2}
  \begin{tabular*}{\columnwidth}{@{}l
    @{\extracolsep{\fill}}c@{\extracolsep{0pt}\hspace{6pt}}c
    @{\extracolsep{\fill}}c@{\extracolsep{0pt}\hspace{6pt}}c
    @{\extracolsep{\fill}}c@{\extracolsep{0pt}\hspace{6pt}}c@{}}
  \toprule
  \textbf{Cat.}
  & \multicolumn{2}{c}{\textbf{$10^\circ$/$10$\,mm  }}
  & \multicolumn{2}{c}{\textbf{$5^\circ$/$10$\,mm  }}
  & \multicolumn{2}{c}{\textbf{$5^\circ$/$5$\,mm  }} \\
  \cmidrule(lr){2-3}\cmidrule(lr){4-5}\cmidrule(lr){6-7}
  & Sim & Real & Sim & Real & Sim & Real \\
  \midrule

  \multicolumn{7}{c}{\textbf{Training Objects}} \\

Plate & 98.8 & 84 & 98.0 & 62 & 93.7 & 30 \\
Disc & 99.8 & 70 & 99.4 & 62 & 98.6 & 10 \\
Rod  & 99.8 & 88 & 99.4 & 60 & 99.4 & 38 \\

\midrule
\textbf{Mean} & 99.5 & 80.7 & 98.9 & 61.3 & 97.2 & 26.0 \\

\midrule
\multicolumn{7}{c}{\textbf{Test Objects}} \\

Plate & 79.5 & 35 & 70.5 & 15 & 44.5 & 15 \\
Disc & 90.0 & 35 & 83.0 & 30 & 77.5 & 15 \\
Rod  & 91.1 & 40 & 85.0 & 20 & 78.3 & 5 \\

\midrule
\textbf{Mean} & 86.9 & 36.7 & 79.5 & 21.7 & 66.8 & 11.7 \\

\bottomrule
\end{tabular*}
\end{table}

 \begin{table}[!htbp]
  \centering
  \caption{\textbf{Refinement performance under the random-target setting.}}
  \label{tab:results_rnd}
  \small
  \renewcommand{\arraystretch}{1.2}
  \begin{tabular*}{\columnwidth}{@{}l
    @{\extracolsep{\fill}}c@{\extracolsep{0pt}\hspace{6pt}}c
    @{\extracolsep{\fill}}c@{\extracolsep{0pt}\hspace{6pt}}c
    @{\extracolsep{\fill}}c@{\extracolsep{0pt}\hspace{6pt}}c@{}}
  \toprule
  \textbf{Cat.}
  & \multicolumn{2}{c}{\textbf{$10^\circ$/$10$\,mm  }}
  & \multicolumn{2}{c}{\textbf{$5^\circ$/$10$\,mm  }}
  & \multicolumn{2}{c}{\textbf{$5^\circ$/$5$\,mm  }} \\
  \cmidrule(lr){2-3}\cmidrule(lr){4-5}\cmidrule(lr){6-7}
  & Sim & Real & Sim & Real & Sim & Real \\
  \midrule

  \multicolumn{7}{c}{\textbf{Training Objects}} \\

  Plate & 78.2 & 72 & 70.4 & 56 & 57.2 & 34 \\
  Disc & 80.8 & 56 & 72.8 & 52 & 64.6 & 16 \\
  Rod  & 98.8 & 50 & 96.4 & 26 & 94.4 & 16 \\

  \midrule
  \textbf{Mean} & 85.9 & 59.3 & 79.9 & 44.7 & 72.1 & 22.0 \\

  \midrule
  \multicolumn{7}{c}{\textbf{Test Objects}} \\

  Plate & 43.0 & 5 & 29.0 & 0 & 15.5 & 0 \\
  Disc & 21.5 & 10 & 16.5 & 5 & 10.0 & 0 \\
  Rod  & 68.1 & 25 & 46.9 & 25 & 33.3 & 15 \\

  \midrule
  \textbf{Mean} & 44.2 & 13.3 & 30.8 & 10.0 & 19.6 & 5.0 \\

  \bottomrule
  \end{tabular*}
  \end{table}

\begin{figure*}[!htbp]
    \centering
    \includegraphics[width=7in]{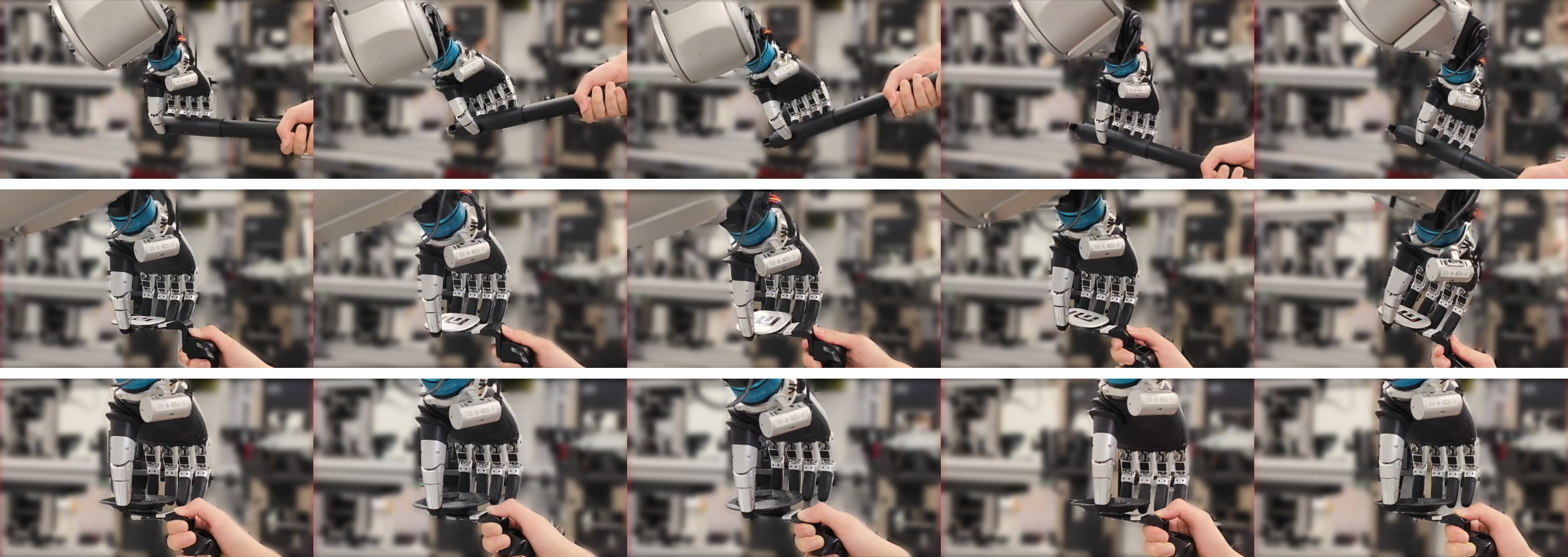}
    \caption{Dynamic tracking of three representative objects by the robotic hand using fingertip tactile feedback.}
    \label{fig:long}
\end{figure*}

\subsection{Main Results}

\subsubsection{Simulation and Real-World Results}
We evaluate two settings: \textbf{(i) fixed-target}, where refinement starts from a random pose toward a fixed tactile target, and \textbf{(ii) random-target}, where both poses are sampled within the dataset range. We run 100 simulation episodes and 10 real-world trials per object.

On training objects, TacRefineNet achieves simulation success rates of 99.5\%, 98.9\%, and 97.2\% under the progressively stricter $10^\circ$/10\,mm, $5^\circ$/10\,mm, and $5^\circ$/5\,mm criteria in the fixed-target setting (Table~\ref{tab:results_fix}). With random targets, the corresponding rates are 85.9\%, 79.9\%, and 72.1\% (Table~\ref{tab:results_rnd}). The decrease is expected because random targets introduce greater variation in both correction direction and magnitude, yet the policy retains 72.1\% success under the strictest criterion.

When transferred to the real system without adaptation, the mean success rates are 80.7\%, 61.3\%, and 26.0\% for fixed targets and 59.3\%, 44.7\%, and 22.0\% for random targets under the same three criteria. The larger degradation under strict thresholds reflects the combined effects of tactile-domain mismatch, sensor noise, and execution error. Under the fixed-target $10^\circ$/10\,mm criterion, rods and plates achieve 88.0\% and 84.0\%, respectively, whereas discs achieve 70.0\%; their rotational symmetry produces less distinctive tactile cues. Under random targets, plates achieve the highest rate of 72.0\%, followed by discs at 56.0\% and rods at 50.0\%.

Fig.~\ref{fig:real_seen} further shows how the real-world errors evolve over the five refinement steps. For both settings, the average position and orientation errors decrease sharply after the first update and then improve more gradually. After five steps, the fixed-target errors reach 5.3\,mm and $3.4^\circ$, while the random-target errors reach 5.1\,mm and $3.5^\circ$. Although the terminal mean errors are similar, random-target success remains lower because the initial configurations and correction trajectories are more diverse, producing greater trial-level variation under the joint position--orientation criteria.

\subsubsection{Generalization to Objects with Unseen Edge Features}

We evaluate two unseen objects per category, each with edge profiles not present during training (Fig.~\ref{fig:objects}). In simulation, the fixed-target success rates are 86.9\%, 79.5\%, and 66.8\% under the three criteria, compared with 44.2\%, 30.8\%, and 19.6\% for random targets. In the real world, these rates decrease to 36.7\%, 21.7\%, and 11.7\% for fixed targets and 13.3\%, 10.0\%, and 5.0\% for random targets (Tables~\ref{tab:results_fix} and~\ref{tab:results_rnd}). This gap relative to the training objects indicates that changes in edge geometry and contact distribution remain a major source of generalization error.

The per-step curves in Fig.~\ref{fig:real_unseen} nevertheless show consistent corrective behavior. The average errors decrease to 6.0\,mm and $5.1^\circ$ for fixed targets and to 6.1\,mm and $4.7^\circ$ for random targets after five steps. However, the larger object-to-object variation and residual errors near the evaluation thresholds lead to substantially lower joint success rates. Under the random-target $10^\circ$/10\,mm criterion, rods retain the best transfer performance at 25.0\%, compared with 10.0\% for discs and 5.0\% for plates, suggesting that their elongated edge contacts provide more transferable local cues. Overall, the policy can reduce pose errors on unseen objects, but reliable refinement still depends on sufficiently discriminative tactile patterns.

\subsection{Ablation Study}
We conduct ablation studies on the input modalities, network components, and the number of tactile fingertips. Unless otherwise specified, results are reported as mean $\pm$ std over three random seeds on seen objects in the random-target simulation setting, with tactile inputs added sequentially from the thumb to the little finger. (As these are three-seed averages of retrained models, the full-model numbers differ slightly from the single deployed checkpoint reported in Table~\ref{tab:results_rnd}.) As summarized in Table~\ref{tab:ablation}, removing tactile input (\textit{W/O Tactile}) causes a dramatic performance drop, increasing the loss from $1.89$ to $100.3$ ($\times10^{-2}$) and reducing success rates under both $10^\circ$/10\,mm and $5^\circ$/5\,mm criteria to nearly zero, confirming that tactile sensing is the primary source of information for grasp refinement. Removing the finger fusion module (\textit{W/O Finger Fusion}) also degrades performance, demonstrating the importance of aggregating tactile information across multiple fingers. Excluding proprioception (\textit{W/O Proprioception}) results in a moderate decrease, indicating that it provides complementary information. Increasing the number of tactile fingertips consistently improves performance, with the largest gain from two to three fingers, while four fingers achieve performance close to the full five-finger model, suggesting that the additional little finger contributes only marginally in our setup.

\begin{table}[htbp]
    \centering
    \caption{\textbf{Ablation study of TacRefineNet.} }
    \label{tab:ablation}
    \small
    \renewcommand{\arraystretch}{1.2}
    \begin{tabular*}{\columnwidth}{@{}l@{\extracolsep{\fill}}ccc@{}}
    \toprule
    \textbf{Variant} & \textbf{Loss ($\times10^{-2}$) } & \textbf{$10^\circ$/$10$\,mm  } & \textbf{$5^\circ$/$5$\,mm  } \\
    \midrule
    \multicolumn{4}{c}{\textbf{Input \& architecture}} \\
    \textbf{Ours (Full)} & $\bm{1.89\pm0.08}$ & $\bm{83.7\pm1.6}$ & $\bm{64.9\pm5.1}$ \\
    W/O Proprioception   & $1.96\pm0.06$     & $79.7\pm0.8$  & $61.6\pm0.6$ \\
    W/O Finger Fusion    & $2.76\pm0.03$     & $73.6\pm0.3$   & $50.4\pm0.9$  \\
    W/O Tactile          & $100.3\pm1.3$ & $0.1\pm0.0$   & $0.0\pm0.0$ \\
    \midrule
    \multicolumn{4}{c}{\textbf{Number of fingers used}} \\
    2 fingers            & $6.56\pm0.30$ & $67.4\pm0.7$ & $45.6\pm1.3$ \\
    3 fingers            & $2.48\pm0.03$ & $82.4\pm0.6$ & $62.2\pm0.7$ \\
    4 fingers            & $2.01\pm0.11$ & $82.0\pm0.2$ & $62.8\pm0.4$ \\
    \bottomrule
    \end{tabular*}
  \end{table}

\begin{table}[!htbp]
  \centering
  \caption{Comparison with Pose-Estimation-Based Baselines.}
  \label{tab:param_compare}
  \small
  \renewcommand{\arraystretch}{1.2}
  \begin{tabular*}{\columnwidth}{@{}l@{\extracolsep{\fill}}ccc@{}}
  \toprule
  \textbf{Method} & \textbf{$10^\circ$/$10$\,mm} & \textbf{$5^\circ$/$10$\,mm} & \textbf{$5^\circ$/$5$\,mm} \\
  \midrule
  \textbf{Ours (direct $\Delta$)} & \textbf{85.9} & \textbf{79.9} & \textbf{72.1} \\
  Baseline (6-DoF Pose)               & 54.1 & 44.1 & 34.7 \\
  \bottomrule
  \end{tabular*}
\end{table}

\subsection{Comparison with the Pose Estimation Baseline}
We compare TacRefineNet with an absolute-pose estimation baseline using the same tactile encoder, training dataset, and evaluation protocol. The baseline independently predicts the absolute 6-DoF wrist poses associated with the current and target tactile observations, and then computes the required wrist increment from their relative SE(3) transformation. In contrast, TacRefineNet jointly processes the paired tactile observations and directly regresses the relative pose increment.

As shown in Table~\ref{tab:param_compare}, our method achieves success rates of 85.9\%, 79.9\%, and 72.1\% under the three thresholds, compared with 54.1\%, 44.1\%, and 34.7\% for the baseline. Absolute pose estimation is ill-conditioned for sparse and geometrically symmetric tactile contacts, and errors from the two independent pose estimates accumulate when computing their relative transformation. Direct increment regression instead focuses on the local difference between the current and target contacts, yielding a better-conditioned learning objective and more accurate corrections.
  
\subsection{Robustness Test}
To evaluate the robustness of the proposed method in dynamic scenarios, we conduct a long-horizon object-tracking experiment using representative objects from the three geometry classes. A fixed tactile image is used as the target observation, while the object pose is continuously perturbed throughout the experiment. The objective is to assess whether the system can continuously adjust the grasp to maintain the desired target configuration. The tactile-based pose refinement process is illustrated in Fig.~\ref{fig:long}, and detailed experimental results are provided in the accompanying video.
As shown in Fig.~\ref{fig:long}, representative tactile-tracking snapshots of the three objects are arranged by rows. The results demonstrate that, even under continuous object-pose variations, the proposed method can stably perform high-precision grasp refinement toward a specified target pose.

\section{Limitation}
Our approach has several limitations. First, it relies on informative tactile patterns and therefore performs best when contact distributions are sufficiently distinctive; objects with symmetric or weakly discriminative contacts, such as discs, remain challenging. Second, the current formulation considers only selected pose dimensions and is designed for local pose adjustments within the sampled dataset range, rather than global grasp planning or large-scale pose correction. Third, although the policy transfers zero-shot from simulation to the real system, the remaining sim-to-real gap still affects performance, especially on unseen objects. Extending the framework to combine tactile refinement with visual perception for global localization and coarse guidance will be an important direction for future work.
\section{Conclusion}
This letter presented TacRefineNet, a goal-conditioned tactile framework for local grasp refinement of edge-prominent objects. The policy uses paired current and target multi-finger tactile observations together with their corresponding hand-joint configurations to predict corrective wrist-pose increments. These corrections are executed through an iterative open--move--regrasp loop, enabling tactile-only feedback control during the final grasping stage. TacRefineNet was trained on 156,007 simulated samples from 15 plates, discs, and rods and transferred to the real hand without real-world fine-tuning. On seen objects, the real system achieved 80.7\% and 59.3\% success for fixed and random targets, respectively, under the $10^\circ$/10\,mm criterion, with mean final errors of approximately 5.2\,mm and $3.5^\circ$. The ablation results confirmed the importance of tactile input and multi-finger fusion, while the robustness and unseen-object experiments showed both continuous corrective behavior and a clear generalization gap. Overall, the results support tactile-guided local refinement within sampled, tactile-observable pose ranges.

\bibliographystyle{IEEEtran}
\bibliography{IEEEabrv, reference}

\end{document}